\documentclass{article}[12pt, oneside, a4paper, times]
\usepackage{graphicx} %allows \includegraphics command to work
\usepackage{float} %allows the [H] tag to stop figures from "floating"
\usepackage{cite}

\newcommand{\be}{\begin{equation} }
\newcommand{\ee}{\end{equation} }
\newcommand{\bss} {\begin{footnotesize}}
\newcommand{\ess} {\end{footnotesize}}
\newcommand{\bea}{\begin{eqnarray}}
\newcommand{\eea}{\end{eqnarray}}

\usepackage[latin1]{inputenc}     % text on ISO-Latin1
\usepackage[T1]{fontenc}     % T1 Font encoding (optional)
\usepackage{latexsym}         % additional math symbols
\usepackage{alltt}        % all-teletype, a better verbatim
\usepackage{fancyhdr}         % better headers
\usepackage{setspace}     % for linespacing
\usepackage{indentfirst}         % first line on every section is indented
\usepackage{epsfig}         % For EPS figures
\usepackage{subfigure}         % The name says it all!!
\usepackage{epic,eepic}         % For EPIC and EEPIC figures
\usepackage{vmargin}         % to change the page formating
\usepackage{amssymb}
\usepackage{amsmath}

\setpapersize [portrait]{A4}
\setmarginsrb {30mm} % margem esquerda
              {10mm} % margem topo
             {15mm} % margem direita
            {20mm} % margem p�
           {2ex}  % altura do espaco para cabe�alho
           {5ex}  % espa�o entre fim do cabe�alho e in�cio do texto
          {0pt}  % altura do espa�o para rodap�
         {20mm}  % espa�o entre fim do texto e fim do rodap�

\begin{document}

\title{A Classifier-free Ensemble Selection Method based on Data Diversity in Random Subspaces \\ Technical Report}
\author{Albert H.R. Ko\\
\'Ecole de technologie sup\'erieure (\'ETS), Universit\'e du Qu\'ebec, \\Montreal, QC, Canada, albert@livia.etsmtl.ca\\
\and
Robert Sabourin\\
\'Ecole de technologie sup\'erieure (\'ETS), 
Universit\'e du Qu\'ebec, \\Montreal, QC, Canada, robert.sabourin@etsmtl.ca\\
\and
Alceu de S. Britto Jr\\
Pontif\'icia Universidade Cat\'olica do Paran\'a (PUCPR), \\Curitiba, PR, Brazil, alceu@ppgia.pucpr.br\\
\and
Luiz E. S. Oliveira\\
Universidade Federal do Paran\'a (UFPR), \\Curitiba, PR, Brazil, lesoliveira@inf.ufpr.br\\
}

\date{}
\maketitle

\begin{abstract}
The Ensemble of Classifiers (EoC) has been shown to be effective in improving the performance of single classifiers by combining their outputs, and one of the most important properties involved in the selection of the best EoC from a pool of classifiers is considered to be classifier diversity. In general, classifier diversity does not occur randomly, but is generated systematically by various ensemble creation methods. By using diverse data subsets to train classifiers, these methods can create diverse classifiers for the EoC. In this work, we propose a scheme to measure data diversity directly from random subspaces, and explore the possibility of using it to select the best data subsets for the construction of the EoC. Our scheme is the first ensemble selection method to be presented in the literature based on the concept of data diversity. Its main advantage over the traditional framework (ensemble creation then selection) is that it obviates the need for classifier training prior to ensemble selection. A single Genetic Algorithm (GA) and a Multi-Objective Genetic Algorithm (MOGA) were evaluated to search for the best solutions for the classifier-free ensemble selection. In both cases, objective functions based on different clustering diversity measures were implemented and tested. All the results obtained with the proposed classifier-free ensemble selection method were compared with the traditional classifier-based ensemble selection using Mean Classifier Error (ME) and Majority Voting Error (MVE). The applicability of the method is tested on UCI machine learning problems and NIST SD19 handwritten numerals.
\end{abstract}


\section{Introduction}
The goal of pattern recognition systems is to achieve the best possible classification performance. Since different classifiers usually make errors on different samples, we can combine classifiers to yield more accurate recognition rates. This approach is known as the Ensemble of Classifiers (EoC) \cite{br-di-2005,  ki-on-1998,  ku-me-2003, op-po-1999, pe-co-2004, we-mu-2004,zo-bu-2004,ki-cl-2008,pa-gr-2008, bri-pr-2014}. Diverse classifiers can be created in several ways, such as Random Subspaces \cite{ho-th-1998, so-ex-2007}, Bagging and Boosting \cite{gr-bo-1998, ku-an-2002, sc-bo-1998}. Random Subspaces creates diverse classifiers by using various feature subsets to train them, while Bagging generates diverse classifiers by randomly selecting subsets of samples to train them. Boosting also uses subsets of samples to train classifiers, but each sample is assigned a probability of being selected: difficult samples have a higher probability of being selected, and easier samples have less chance of being used for training. All these ensemble generation methods take advantage of diverse data, which are assembled by extracting only a subset of features or a subset of samples, thereby creating diverse classifiers.

In general, the classifiers created are stored in a pool of classifiers; however, not all the classifiers in this pool will be useful. To select the most pertinent classifiers from the pool \cite{ba-an-2003, br-di-2005, ko-bi-1996, ku-me-2003, pa-so-1997, ru-cl-2005, ue-ge-1996}, we need to define an adequate objective function. This objective function can be a fusion function, like the majority voting error \cite{br-di-2005, ku-me-2003, pa-so-1997, ru-cl-2005}, or simply the diversity among classifiers \cite{gi-dy-2001, me-cr-2005}.

The two key issues that are crucial to the success of an EoC are the following: first, we need diversity for ensemble creation, because an EoC will not perform well without it \cite{ki-on-1998, ku-an-2002, ku-me-2003, ru-an-2001, ru-cl-2005, pa-bi-2009}; and second, we need to select classifiers once they have been created \cite{br-di-2005, ku-an-2002, ku-me-2003, ru-cl-2005, ul-in-2009}, because not all the classifiers created are useful. However, the classical framework - ensemble creation followed by ensemble selection - has some disadvantages. One of these is additional classifier training. Since some of the classifiers created will not be used, time is wasted in training them. Another is the evaluation of high dimensional classifier combinations,  because of the need to evaluate different combinations for ensemble selection following classifier training, which is very time-consuming for a large classifier pool. Hence, our question: Can we select data subsets for ensemble creation directly, instead of performing the ensemble creation/ensemble selection routine?

To answer this question, we assume that data subset selection might be feasible through the evaluation of the diversity of the data in the subsets. This means that, by clustering the data points in different feature subspaces, we might have quite diverse clustering partitions. Since clustering diversities measure the diversity of these partitions, they give an indirect indication of the data diversity of the feature subspaces. From this assumption, we use clustering diversity to represent the data diversity of different feature subsets in Random Subspaces. Thus, the use of clustering diversity as the data diversity measure could allow us to apply a classifier-free ensemble selection scheme.

With this scheme, it is only necessary to train one classifier for each feature subset selected to evaluate the performances of ensembles, so that the best ensemble can be chosen. Compared with a traditional ensemble selection scheme, which requires the training of all classifiers and combinations of all the ensembles evaluated, the proposed scheme offers an interesting alternative for tackling problems with a large classifier pool and time-consuming classifier training. This classifier-free method is only for use with the Random Subspaces ensemble generation method. Remember that different classifiers are generated with all samples, but only some of the features are used in the Random Subspaces method. Since we generate different classifiers based on different feature subsets, then, if we can select diverse feature subsets, we are actually selecting
adequate classifiers. We thus propose a method for feature subset selection on Random Subspaces which will also constitute a classifier-free ensemble
selection method. With this approach, we can reduce the time spent on useless classifier training and also reduce the ensemble selection search space.

    \begin{figure}[!htb]
      \begin{center}
      \includegraphics[width=.9\textwidth]{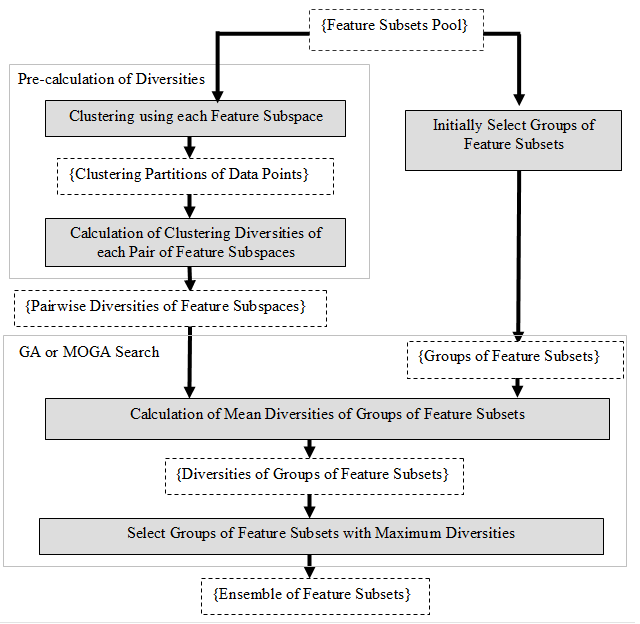}
      \caption{The proposed classifier-free ensemble selection scheme is, in fact, a feature subset selection in Random Subspaces. We carried out this feature subset selection using clustering diversity as the objective function. Note that the pre-calculation of diversities is carried out once and for all, while the GA or MOGA search is repeated from generation to generation.}
\label{fs}
      \end{center}
    \end{figure}

Here, we need to clarify the concept of clustering diversity. In general, it is meant to help in the construction of a cluster ensemble, and has nothing to do with classifiers. A cluster ensemble combines the results of several partitions and thus improves the quality and robustness of partitions of data \cite{di-vo-2001, fe-ra-2003, fi-ba-2003, fr-da-2002, ku-us-2004,pa-di-2003,  qi-cl-2000,  st-cl-2002, to-co-2003,  to-cl}. It has been shown that more diverse cluster ensembles offer the potential for greater improvement than do less diverse cluster ensembles \cite{fe-ra-2003}, and that is why we use clustering diversity in our study.

Given a pool of feature subsets, we use a clustering algorithm with fixed parameters to form clusterings in feature subsets (Fig. \ref{fs}). It is reasonable to assume that if we ensure clustering diversity between different feature subsets, then we also ensure data diversity.

\begin{enumerate}
\item By selecting the useful feature subsets, we can reduce the time needed for classifier training for ensemble creation.
\item By evaluating the pertinent feature subsets, we can significantly reduce the search space for ensemble selection.
\item Feature subset selection might be able to replace ensemble selection completely for Random Subspaces in some circumstances, and this would enable it to offer de facto classifier-free ensemble selection.
\end{enumerate}

Our experimental results suggest that there is a strong correlation between classifier diversity and clustering diversity in Random Subspaces, and that clustering diversity does work for a classifier-free ensemble selection scheme.  Here, we need to mention that the proposed strategy would not work for the Bagging and Boosting ensemble generation methods. Since both Bagging and Boosting draw a certain proportion of the data points to train classifiers, it is quite possible that the distributions of data points will be rather similar. Consequently, clustering these data points might not generate significantly different clustering partitions. More importantly, since Bagging uses various data points for each classifier, it is impossible for us to measure data diversity by clustering different parts of data points.

In the next section, we introduce general clustering diversity measures. In section $3$, we investigate the possibility of ensemble selection using clustering diversity measures on the UCI Machine Learning Repository. Moreover, we report the experiments we performed on NIST SD19 handwritten numeral digits. Discussion is provided in section $4$, and our conclusion is presented in the last section.

 \section{Clustering Diversity Measures}
In general, given two clustering partitions, we can apply clustering diversity to measure the diversity between them. Since there is no class label available in clustering, the concept of diversity based on correct/incorrect classification cannot be applicable for clustering diversity, and another kind of approach will be needed. First, we introduce the concept of clustering diversity from the framework defined in \cite{me-co-2002}. For $C$ data points, if we suppose that one clustering, $C_{i}$, groups these data points into $I$ clusters, and another clustering, $C_{{k}}$, groups them into $K$ clusters, then the diversity between these two clusterings can be deduced as follows:
\subsection{Basic Concept of Clustering Diversity}
For two clusterings, consider a contingency table (or confusion matrix) $M$ as a $I\times K$ matrix which describes the partitions of data points in these two clusterings. If we look at the ${ik}_{th},1\le i \le I, 1 \le k \le K$ element in the contingency table $M$ - let us call it block $M_{{ik}}$ - which represents data points grouped as a cluster by clustering $C_{i}$ and also groups data points as a cluster by clustering $C_{k}$, we find that all the data points that are grouped into cluster $c_{{i}}$ by clustering $C_{i}$ and grouped into cluster $c_{{k}}$ by clustering $C_{k}$ are located in the block $M_{{ik}}$. So, in this contingency table, we can denote the number of data points in block $M_{{ik}}$ as $ m_{{ik}}$:
 \bss
 \bea
 m_{{ik}} = |c_{i} \bigcap c_{k}| \\
   \sum_{1\le i \le I}  \sum_{1 \le k \le K} m_{ik} = C \label{equ6-2}
 \eea
\ess
We note that, given two clusterings, the complexity of the calculation of all $ m_{{ik}}$ is $O(C \cdot (I+K))$. Once we have every element $ m_{{ik}}$ for contingency table $M$, we can use $ m_{{ik}}$ to calculate the clustering diversity between clustering $C_{i}$ and clustering $C_{k}$. Given that we have $C$ data points, we want to determine the relationships between these $\frac {C\cdot (C-1)} {2}$ data point pairs. We then classify these relationships into four different cases and count the occurrences of these cases:
   \begin{enumerate}
 \item $C_{11}$: \bigskip the number of data point pairs in the same cluster under both $C_{i}$ and $C_{k}$

 \item $C_{00}$: \bigskip the number of data point pairs in different clusters under both $C_{i}$ and $C_{k}$

 \item $C_{10}$: \bigskip the number of data point pairs in the same cluster under $C_{i}$, but not under $C_{k}$

 \item $C_{01}$: \bigskip the number of data point pairs in the same cluster under $C_{k}$, but not under $C_{i}$
 \end{enumerate}
If we suppose that we have $C$ points in total, then the following condition must be satisfied:
\bss
\be
C_{11} + C_{00} + C_{10} + C_{01} = \frac {C (C-1)} {2}
\ee
\ess

\begin{figure}[!htb]
      \begin{center}
      \includegraphics[width=.8\textwidth]{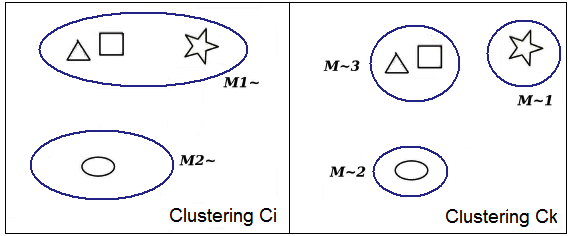}
      \caption{Illustration of  $2$ clustering partitions. The first clustering generates $2$ partitions and the second clustering generates $3$ partitions.}
\label{mik_clus}
      \end{center}
    \end{figure}

\begin{figure}[!htb]
      \begin{center}
      \includegraphics[width=.6\textwidth]{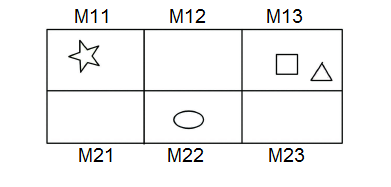}
      \caption{The $2$ partitions of the first clustering can be denoted as ($M_{1k}$ and $M_{2k}$), and those of the second clustering can be denoted as ($M_{i1}$,  $M_{i2}$, and $M_{i3}$). All data points are classified into $M_{ik}$ based on these partitions.}
\label{Mik}
      \end{center}
    \end{figure}

\begin{figure}[!htb]
      \begin{center}
      \includegraphics[width=.7\textwidth]{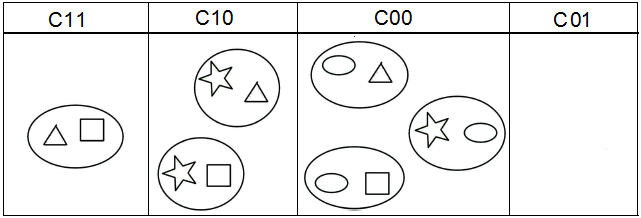}
      \caption{Examples of the calculation of $C_{11}, C_{10}, C_{00}$, and $C_{01}$ based on $4$ data points, and thus on $6$ data point pairs.}
\label{Cik}
      \end{center}
    \end{figure}

To illustrate the meanings of $C_{ij}$ in Fig. \ref{mik_clus} and Fig. \ref{Mik}, we carried out $2$ clusterings on $4$ data points. Note that these $4$ data points mean $6$ data point pairs. In Fig. \ref{Cik}, $C_{11}=1$, because the triangle and the rectangle are grouped together in the same clusters by both clusterings. $C_{10}=2$, because the star is grouped in the same cluster as the triangle and the rectangle by one clustering, but into different clusters by another clustering. By a similar analysis, we can observe that $C_{01}=0$.  $C_{00}=3$, because the ellipse is considered to be in a different cluster from the star, the triangle, and the rectangle by both clusterings.

While the direct calculation of $C_{11}, C_{00}, C_{10}$, and $C_{01}$ could be very time-consuming - the complexity is $O(\frac {C (C-1)} {2})$ - this calculation can be greatly accelerated. In fact, all the values $C_{11}, C_{00}, C_{10}$, and $C_{01}$ can be quickly derived from the contingency table $M$ using its element $m_{{ik}}$.

If we suppose there are $m_{ik}$ data points in block $M_{{ik}}$, then we can calculate the $C_{11}$ value as the data point pairs in this block, i.e. $C_{11} (M_{ik}) = \frac {m_{ik} (m_{ik}-1)} {2}$. Consequently, the total $C_{11}$ value can be calculated as the sum of $C_{11} (M_{ik})$ from all these blocks, i.e. $C_{11} =  \sum_{1\le i \le I}  \sum_{1 \le k \le K} C_{11} (M_{ik})$:

\bss
\be
C_{11} =  \sum_{1\le i \le I}  \sum_{1 \le k \le K} \frac {m_{ik} (m_{ik}-1)} {2} \\
\ee
\ess
Using eq. \ref{equ6-2}, we can write:
\bss
\be
C_{11} = \frac {(   \sum_{1\le i \le I}  \sum_{1 \le k \le K} m_{ik}^{2} ) - C} {2} \label{equ6-5}
\ee
\ess
For $C_{10}$ and $C_{01}$, the calculation follows the same principle. It can be deduced that there are $\frac {((\sum_{i} m_{ik}) - m_{ik})} {2}$ data point pairs grouped in the same cluster by clustering $C_{i}$, but in different clusters by clustering $C_{k}$. Consequently, we can arrive at a value for $C_{10}$:
 \bss
\be
C_{10} = \frac {   \sum_{1\le i \le I}  \sum_{1 \le k \le K} m_{ik} ( (\sum_{i} m_{ik}) - m_{ik})} {2} \label{equ6-6}
\ee
\ess
For $C_{01}$, we can use the same method and obtain a similar result:
\bss
\be
C_{01} = \frac {   \sum_{1\le i \le I}  \sum_{1 \le k \le K} m_{ik} ((\sum_{k} m_{ik}) - m_{ik})} {2} \label{equ6-7}
\ee
\ess

The more complicated case is the deduction of $C_{00}$, for which we should look for data point pairs grouped in different clusters by both the $C_{i}$ and the $C_{k}$ clustering. Since there are $(C - \sum_{k} m_{ik}- \sum_{i} m_{ik} + m_{ik})$ samples satisfying this condition, we can arrive at the following equation:

\bss
\be
C_{00} = \frac {   \sum_{1\le i \le I}  \sum_{1 \le k \le K} (m_{ik} \cdot (C - \sum_{k} m_{ik}- \sum_{i} m_{ik} + m_{ik}))} {2} \label{equ6-8}
\ee
\ess
The result can be verified by calculating $C_{11} + C_{10} + C_{01} + C_{00} = \frac {C (C-1)} {2}$.

Remember that the complexity of the calculation of all $ m_{{ik}}$ is $O(C \cdot (I+K))$. Given that $I, K \ll C$, the calculation of $C_{11}, C_{00}, C_{10}$, and $C_{01}$ deduced by $ m_{{ik}}$ is much faster than the direct calculation of $C_{11}, C_{00}, C_{10}$, and $C_{01}$, which had a complexity of $O(\frac {C (C-1)} {2})$.

We need to mention that we fix all the clustering parameters, including the number of clusters. In other words, in our case, $I = K$, and the contingency table $M$ is, in fact, a square matrix.

However, these four types of relationships of data point pairs are not themselves clustering diversity measures. In fact, several different clustering diversity measures have been proposed using the counts of these four cases. We introduce them in the next section.

\subsection{Pairwise Clustering Diversity Measures}
Based on the pairwise counts, a number of clustering diversity measures are proposed \cite{me-co-2002}:

 \begin{enumerate}
 \item  Wallace Indices\\
 \bss
\bea
Wallace-1:W_{i}(C_{i}, C_{k})= \frac {C_{11}} {C_{11}+ C_{10}} \\
Wallace-2:W_{k}(C_{i}, C_{k})= \frac {C_{11}} {C_{11}+ C_{01}}
\eea
 \ess
 \item Fowlkes-Mallows Index\\
 \bss
\be
 F(C_{i}, C_{k})= \frac {C_{11}} { ((C_{11}+C_{10})(C_{11}+C_{01}))^{ \frac {1} {2}} } = (W_{i}(C_{i}, C_{k})W_{k}(C_{i}, C_{k}))^{\frac {1} {2}}
\ee
\ess
 \item Rand Index \\
 \bss
\be
 R(C_{i}, C_{k}) = \frac {C_{11}+C_{00}} {\frac {C(C-1)} {2}}
\ee
\ess
 \item Jacard Index \\
\bss
\be
 J(C_{i}, C_{k}) = \frac {C_{11}} {C_{11}+C_{01}+C_{10}}
\ee
\ess
 \item Mirkin's Metric \\
\bss
\be
 K(C_{i}, C_{k}) = 2(C_{10} + C_{01}) = C(C-1)[1 -  R(C_{i}, C_{k})]
\ee
\ess
 \end{enumerate}
Note that all these measures calculate the clustering diversity between two clusterings. In the case where there are more than two clusterings, the global clustering diversity is simply the mean of all clustering diversities between all clustering pairs. Given $L$ clusterings, there are $\frac {L \times (L-1)} {2}$ clustering diversities $d_{12}, d_{13}, ..., d_{(L-1)L}$ to be calculated, and the global clustering diversity $\bar{d}$ will be its average:
\bss
\be
{\small \bar{d} } = 2 \times \frac {\sum_{ij} d_{ij}} {L \times (L-1)}, i < j  \label{equ6-15}
\ee
\ess

At this point, we wanted to check whether or not the clustering diversity of different feature subsets could be used as an objective function for classifier-free ensemble selection, and so we carried out experiments on UCI machine learning problems (see below).

\section{Experiments}
\begin{figure}[!htb]
      \begin{center}
      \includegraphics[width=.55\textwidth]{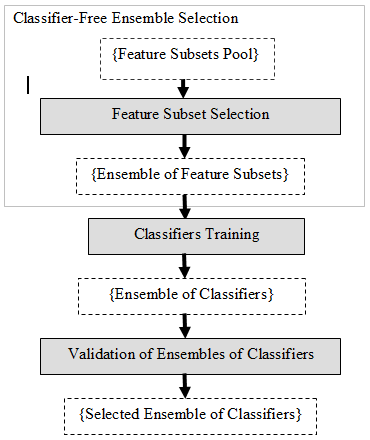}
      \caption{The processing steps of the proposed classifier-free ensemble selection method. The selected ensembles of feature subsets can be used to train ensembles of classifiers, which must then be tested in a validation set in order to select the best one. The details of feature subset selection are shown in Fig. \ref{fs}.}
\label{plan}
      \end{center}
    \end{figure}

This section describes the experiments undertaken to show the applicability of the proposed method. First, we needed to evaluate the hypothesis that the clustering diversity of different feature subsets could be used as an objective function for ensemble selection in Random Subspaces. For an ensemble created with the Random Subspaces method, we first evaluated its feature subspaces by carrying out simple K-Means clusterings with predefined numbers of clusters on these feature subsets. The number of clusters is preselected using the Xie-Beni index (XB index) \cite{ba-no-2001, ha-on-2001} as the clustering validity index.  A clustering diversity was thus calculated based on the clusterings of these feature subsets, and served as an objective function for the search. Six clustering diversities were tested in our experiment, including: Mirkin's Metric, two Wallace Indices, the Fowlkes-Mallows Index, the Rand Index and the Jacard Index. As we mentioned in the introduction, the search algorithm is also an important issue for ensemble selection. For the classifier-free ensemble selection scheme, we evaluate two types of search algorithms: the single genetic algorithm (GA) and the multi-objective genetic algorithm (MOGA). We used the GA because, as a population-based search algorithm, it is flexible and its complexity can be adjusted according to the size of the population and the number of generations. Moreover, because the algorithm returns a population of the best combinations, it can be potentially exploited to prevent generalization problems \cite{ru-cl-2005}. Once the feature subsets had been selected, we constructed corresponding classifiers using the selected feature subsets and evaluated the performance of the ensembles of these classifiers (see Fig. \ref{plan}).

At the same time, we needed to compare our classifier-free ensemble selection scheme to traditional classifier-based ensemble selection methods. In traditional classifier-based ensemble selection, each feature subset is used to train a classifier, and all the trained classifiers are stored in a pool. In order to select adequate classifiers from this pool, we carried out the ensemble selection process using Majority Voting Error (MVE) as the objective function for the GA and MOGA search algorithms.

In our experiments, both the GA and the MOGA are based on bit representation, one-point crossover, bit-flip mutation, roulette wheel selection, and elitism implemented using a generational procedure. Population sizes ranging from 16 to 256 were assessed and the best trade-off performance/computational cost was achieved using 32 individuals in the population.

The evaluation of objective functions for ensemble selection is tested on UCI machine learning problems and also on a large-scale problem by using the NIST SD19 handwritten numerals. 

\subsection{Evaluation on the UCI Machine Learning Repository}

We performed the classifier-free ensemble selection and classifier-based ensemble selection experiments on UCI machine learning problems (Table \ref{t29}). Three classification algorithms were used for the classification tasks: Quadratic Discriminant Classifiers QDC), K-Nearest Neighbors Classifiers (KNN), and Parzen Windows Classifiers (PWC) \cite{du-pa}.

\begin{table}
\begin{scriptsize}
\begin{center}
 \begin{tabular}{|c|c|c|c|c|c|c|}
  \hline
  database &  number of  & number of & number of & number of & number of & number of \\
  	 &  classes  & clusters & train samples & test samples & features & cardinality \\
 \hline \hline
  Pima-Diabetes  			& 2  & 3   & 384   & 384  & 8  & 4  \\
 Liver-Disorders  			& 2  & 5   & 144   & 144  & 6  & 3  \\
 Wisconsin Breast-Cancer 	& 2  & 12  & 284   & 284  & 30 & 5  \\
  Wine					& 3  & 4   & 88    & 88   & 13 & 6  \\
 Image Segmentation 		& 7  & 53  & 210   & 2100 & 19 & 4  \\
 Letters Recognition 		& 26 & 87 & 10000 & 10000 & 16 & 12 \\
 \hline
 \end{tabular}
\caption{The problems extracted from the UCI Machine Learning Repository.}  \label{t29}
\end{center}
\end{scriptsize}
\end{table}

\begin{figure}[!htb]
      \begin{center}
      \includegraphics[width=.90\textwidth]{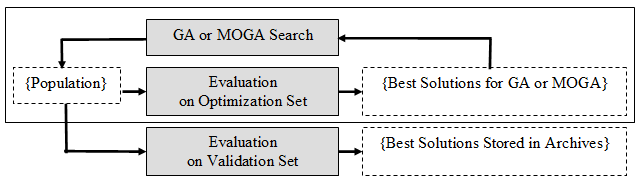}
      \caption{The archive validation set is used to validate the population found by a GA or a MOGA search and then stores the best solutions in a separate archive.}
\label{arch}
      \end{center}
    \end{figure}

All the problems extracted from the UCI data repository have two datasets, a training set for classifier training for the GA or MOGA search and a test set used only for testing. The whole training set was used to create $10$ classifiers in Random Subspaces. Moreover, the training samples were divided into $3$ parts for each scheme:

\begin{itemize}
\item Optimization set: \\
$70 \%$ of the training samples were used for both the GA and the MOGA search. These samples were clustered in feature subspaces, and the clustering diversity indices were measured by comparing clusterings in a pairwise manner. The diversity of a set of feature subspaces is calculated as the mean value of pairwise diversities of the features involved (eq. \ref{equ6-15}).

\item Archive validation set: \\
Another $15 \%$ of the training samples were used as the archive validation mechanism \cite{ra-an-2006} to avoid overfitting during the GA or MOGA search. They were used to evaluate all the individuals and then to store the optimal solutions in a separate archive after each generation (Fig. \ref{arch}). The reason for using this archive validation mechanism is that solutions found in a Pareto front of one dataset may be optimal only for this special search dataset. From generation to generation, the solutions found may tend to overfit the search dataset. To make sure that those solutions were not overfitted in our case, we validated them in another archive validation set. The solutions are stored in the archive only if they dominate all the solutions in that set.

\item Classifier-free MOGA evaluation set: \\
The remaining $15 \%$ of the training samples were used solely for the final classification performance validation for the classifier-free MOGA search. The reason for this was that, unlike the GA search, which gives the best individual in the population, a MOGA search gives a group of individuals, called a Pareto front. As a result, we need a way to evaluate the solutions found in this Pareto front. Even though a MOGA search is a purely classifier-free process, the evaluation of these potential solutions will require the construction of classifiers. So, during this process, the feature subset candidates stored in the archive are then used to construct ensembles and their performances evaluated on these samples.

\item Test set: \\
The best solutions found were evaluated on the test set.
\end{itemize}

The classifier-free GA search used the clustering diversities calculated from the optimization set to search for feature subspaces with the maximum clustering diversity. During the search, solutions found in each generation were evaluated with clustering diversity in the archive validation set and stored in an archive. Finally, solutions stored in the archive were used on a test set.

The classifier-free MOGA search follows the same procedure as the classifier-free GA search, except that the former has two objective functions: maximization of clustering diversity and maximization of the number of feature subspaces. We will discuss in the next section the reason for maximizing the number of feature subspaces. Moreover, since the classifier-free MOGA search provides a group of solutions instead of one solution as in the classifier-free GA search, we needed to evaluate the solutions stored in the archive. We trained EoCs using subspaces found by the classifier-free MOGA search, and then evaluated them in a classifier-free MOGA evaluation set, using the best ensemble on a test set.

The classifier-based GA search first constructed all the classifiers using the training set, and then used mean ME (Mean Classifier Error) or MVE evaluated on the optimization set to search for EoCs with the ME or MVE. Again, during the search, solutions found in each generation were evaluated in the archive validation set and stored in an archive. Finally, solutions stored in the archive were used on a test set.

The classifier-based MOGA search also constructed all the classifiers using the training set, and then used the ME or MVE evaluated on the optimization set to search for EoCs with the ME or MVE. However, in order to compare this search with the classifier-free MOGA search, it also used the maximization of the number of feature subspaces as another objective function. Following the MOGA search, the best solutions were selected as the individual at the Pareto front with the minimum error rate. These solutions were then used on a test set. Because the error rate had already been evaluated during the search, the classifier-based MOGA search did not need to use an external evaluation set for the final evaluation, as was done in the case of the classifier-free MOGA search.

We first carried out the experiments with a single GA search, and then we compared the results with those of a MOGA search. 

\subsubsection{Search with Single GA for the UCI Machine Learning Problems}

%\subsection{Search with the Single Genetic Algorithm}

For classifier-free ensemble selection (or feature subset selection), we used different clustering diversity indices as objective functions to find the potentially diverse feature subsets. Among these objective functions, we minimized two Wallace indices, the Fowlkes-Mallows index, the Rand index, the Jacard index, and the maximized Mirkin Metric. All the global clustering diversity measures are calculated as the mean values of clustering diversities between all the clustering pairs. Note that the clustering diversity between any two clustering pairs can be calculated prior to the GA search, so that during the GA search we simply calculate the mean of the clustering diversities among the selected clusterings. For each of the $6$ problems extracted from the UCI data repository, $10$ feature subsets with fixed cardinality are given as the pool for the search. Table \ref{t29} shows the number of features and the cardinality of each database, and also the number of classes, clusters, training and testing samples.
Using the pre-calculated clustering diversities based on the clusterings with these feature subsets, the GA search evaluated the global diversity of various combinations of these feature subsets. The combination with the best global diversity was regarded as the best solution, and then the selected feature subsets were used to construct the classifiers needed. These classifiers were then combined using the Majority Voting (MAJ) fusion function, which is based on a Simple Majority Voting Rule, to give the classification results. The MAJ fusion function does not require the a posteriori outputs for each class, and each classifier gives only one crisp class output as a vote for that class. Then, the ensemble output is assigned to the class with the maximum number of votes among all classes.

\begin{table}
\begin{scriptsize}
\begin{center}
  \begin{tabular}{|c|c|c|c|c|c|}
   \hline
   \hline
  &  Mirkin's & Wallace Index-1     & Wallace  Index-2     &  Fowlkes-Mallows  &    Rand  \\
       \hline
         \hline
  Pima-Diabetes  &79.77 $\pm$1.73  \%  &76.61 $\pm$ 1.74 \%  &77.37 $\pm$1.85  \%  &78.32 $\pm$ 2.59 \%  &77.22  $\pm$ 2.85 \%    \\
Liver-Disorders &  72.11 $\pm$ 2.45 \%  &70.35 $\pm$ 3.49 \%  &72.01 $\pm$ 3.06 \%  &70.39 $\pm$ 4.33 \%  &69.00  $\pm$ 3.68  \%  \\
Wisconsin Breast-Cancer  &  92.18 $\pm$ 0.70 \%  & 89.19 $\pm$ 4.77 \%  &89.71 $\pm$ 4.21 \%  &89.67 $\pm$ 4.71 \%  &91.73  $\pm$ 0.84 \%   \\
Wine  &  75.61 $\pm$ 5.71 \%  &73.52 $\pm$ 1.98 \%  & 73.60 $\pm$ 2.58 \%  &74.05 $\pm$ 3.70 \%  &71.82 $\pm$ 4.71  \%   \\
Image Segmentation  &  74.78 $\pm$ 2.31 \%  &76.87 $\pm$ 3.63 \%  &77.29 $\pm$ 2.96 \%  &78.28 $\pm$ 2.10 \%  &75.29$\pm$1.79  \%  \\
Letters Recognition  &  82.17 $\pm$0.85  \%  &76.48 $\pm$ 3.36 \%  &78.11 $\pm$ 3.90 \%  &77.12 $\pm$ 4.33 \%  &77.85  $\pm$3.35  \%    \\
 \hline
 \hline
 &  Jacard & M.V.E & M.E. & ALL  & Oracle \\
    \hline
         \hline
Pima-Diabetes &  81.35 $\pm$ 1.64 \%  &79.85 $\pm$ 2.36 \% (3.97) &79.57 $\pm$ 2.20 \% (3.83) &  {\bf 82.55}  $\pm$ 0.00 \% & 98.18 \% \\
Liver-Disorders  & 72.11 $\pm$ 2.94 \%  &73.91 $\pm$ 2.89 \% (4.07) &72.29 $\pm$ 2.73 \%  (3.67) & {\bf 76.39 }  $\pm$ 0.00 \% & 100.00 \% \\
Wisconsin Breast-Cancer  &  91.97 $\pm$ 3.69  \%  &92.10 $\pm$ 1.98 \% (3.73) &92.55    $\pm$ 0.85 \% (4.20) &  {\bf 92.61}  $\pm$ 0.00 \%  &  99.65 \% \\
Wine &  72.42 $\pm$ 2.29 \%  &72.50 $\pm$ 1.39 \% (3.63) &75.00 $\pm$ 3.54 \% (3.93) & {\bf  76.14  }  $\pm$ 0.00 \% & 97.73 \% \\
Image Segmentation  &   78.47 $\pm$ 2.68 \%  &72.85 $\pm$ 1.42 \% (4.03) &75.33 $\pm$ 4.21 \% (3.97) & {\bf 78.19 }  $\pm$ 0.00 \%  & 97.29 \% \\
Letters Recognition   & 76.37 $\pm$  3.80\%  &79.99 $\pm$  2.27\% (4.37) &79.25 $\pm$  3.00\% (3.90) & {\bf 83.08 }  $\pm$ 0.00 \% &  94.78 \% \\
\hline
\hline
  \end{tabular}
\caption{The average recognition rates of KNN classifiers selected by a GA with different objective functions. The MVE and ME average ensemble sizes are shown in parentheses.}   \label{t30}
\end{center}
\end{scriptsize}
\end{table}

\begin{table}
\begin{scriptsize}
\begin{center}
  \begin{tabular}{|c|c|c|c|c|c|}
   \hline
   \hline
  &  Mirkin's & Wallace  Index-1     & Wallace  Index-2     &  Fowlkes-Mallows   &    Rand   \\
       \hline
         \hline
Pima-Diabetes  & 76.05 $\pm$ 1.53 \%  &72.74 $\pm$ 2.56 \%  &74.84 $\pm$ 4.16 \%  &74.00 $\pm$ 2.80 \%  &72.86 $\pm$ 3.00 \%   \\
Liver-Disorders &   59.51 $\pm$ 0.45 \%  &57.11 $\pm$ 2.67 \%  &58.12 $\pm$ 2.54 \%  &57.15 $\pm$ 3.34 \%  &59.91 $\pm$ 1.48 \%  \\
Wisconsin Breast-Cancer  &   95.21 $\pm$ 1.11 \%  &91.50 $\pm$ 2.03 \%  &92.50 $\pm$ 2.23 \%  &91.54 $\pm$ 1.15 \%  &93.22 $\pm$ 1.94 \%   \\
Wine  &   95.45 $\pm$ 1.08 \%  &95.76 $\pm$1.26  \%  &93.98 $\pm$ 2.82 \%  &92.73 $\pm$ 3.55 \%  &92.84  $\pm$ 3.75 \%   \\
Image Segmentation  &  72.03 $\pm$ 15.40 \%  &69.85 $\pm$ 13.19 \%  &67.59 $\pm$ 15.43 \%  &74.34 $\pm$ 9.29 \%  &72.89 $\pm$ 12.09 \%   \\
Letters Recognition  &  82.53 $\pm$ 0.97 \%  &82.71 $\pm$ 1.03 \%  &82.36 $\pm$ 1.11 \%  &82.57 $\pm$ 1.50 \%  &82.71 $\pm$  0.88\%   \\
      \hline
         \hline
&  Jacard & M.V.E & M.E. & ALL   &Oracle  \\
    \hline
         \hline
Pima-Diabetes &   75.92 $\pm$ 1.60 \%  &75.49 $\pm$ 2.46 \% (4.30) &74.34    $\pm$ 2.65 \% (3.83) & {\bf 77.86 } $\pm$  0.00 \%  &  93.23 \%\\
Liver-Disorders &   58.63 $\pm$ 2.01 \%  &57.15 $\pm$  2.26\% (4.23) &56.99   $\pm$ 2.70 \% (4.17) & {\bf 57.64 } $\pm$   0.00\% &  88.19 \%\\
Wisconsin Breast-Cancer &  91.55 $\pm$ 1.40 \%  &93.57 $\pm$ 2.06 \% (3.80) &93.69   $\pm$ 1.48 \% (4.07) &{\bf  93.66 } $\pm$ 0.00  \%  &  99.65 \% \\
Wine  & 93.30 $\pm$ 3.71 \%  &92.61 $\pm$ 1.75 \% (4.43) &95.00  $\pm$ 2.44 \% (4.00) & {\bf  96.59} $\pm$  0.00 \%& 100.00 \%\\
Image Segmentation  & 73.23 $\pm$ 12.31 \%  &60.59 $\pm$  12.92\% (3.80) &57.27 $\pm$ 15.65 \% (4.20) & {\bf 78.24 } $\pm$   0.00 \%&  95.29 \%\\
Letters Recognition  & 82.46 $\pm$ 1.52 \%  &81.13 $\pm$  2.37\% (3.80) &84.10  $\pm$ 0.00 \% (9.00) & {\bf 84.36 } $\pm$  0.00 \%& 93.40 \% \\
\hline
  \end{tabular}
\caption{The average recognition rates of QDC classifiers selected by a GA with different objective functions. The MVE and ME average ensemble sizes are shown in parentheses.}   \label{t31}
\end{center}
\end{scriptsize}
\end{table}

\begin{table}
\begin{scriptsize}
\begin{center}
  \begin{tabular}{|c|c|c|c|c|c|}
   \hline
   \hline
  &  Mirkin's & Wallace Index-1     & Wallace  Index-2     &  Fowlkes-Mallows  &    Rand  \\
       \hline
         \hline
  Pima-Diabetes  & {\bf  78.28 }$\pm$ 1.52 \%  &73.87 $\pm$ 2.94 \%  &77.87 $\pm$ 2.56 \%  &76.22 $\pm$ 3.67  \%  &75.44  $\pm$ 3.16 \%  \\
Liver-Disorders &  70.02 $\pm$2.06  \%  &61.34 $\pm$ 2.95 \%  &63.54 $\pm$ 4.06 \%  &62.85 $\pm$ 5.17 \%  &68.12 $\pm$ 3.30 \%    \\
Wisconsin Breast-Cancer  &   90.77 $\pm$ 1.14 \%  &90.16 $\pm$ 1.12  \%  &89.51 $\pm$ 1.51 \%  &90.18 $\pm$1.48  \%  &90.96 $\pm$ 0.31  \%   \\
Wine  & {\bf   81.40} $\pm$ 4.89 \%  &76.74 $\pm$ 2.31 \%  &75.80 $\pm$ 3.06 \%  &76.63 $\pm$ 3.79 \%  &75.72 $\pm$ 5.32 \%   \\
Image Segmentation  &  74.91 $\pm$ 4.20 \%  &72.68 $\pm$ 7.67 \%  &76.89 $\pm$ 2.68 \%  &76.73 $\pm$ 5.98 \%  &72.51$\pm$  7.72\%   \\
Letters Recognition  &   89.00 $\pm$ 0.52 \%  &88.46 $\pm$ 1.05 \%  &88.23 $\pm$ 1.01 \%  &88.37 $\pm$ 1.26 \%  &88.54 $\pm$0.76  \%   \\
 \hline
 \hline
 &  Jacard & M.V.E & M.E. & ALL  & Oracle \\
    \hline
         \hline
Pima-Diabetes &  78.31 $\pm$ 1.75 \%  &77.74 $\pm$ 2.21 \% (4.13)  &78.19    $\pm$ 1.88 \%  (4.03)   & 78.12   $\pm$ 0.00 \%   & 92.19 \%\\
Liver-Disorders  &  63.06 $\pm$ 4.94 \%  &66.76 $\pm$4.07  \% (3.80) &67.87  $\pm$  3.77\% (4.07) &  {\bf 70.83}  $\pm$ 0.00 \%& 89.58 \%\\
Wisconsin Breast-Cancer  &   90.85 $\pm$ 1.18 \%  &90.99 $\pm$ 1.39 \% (4.10) &87.88   $\pm$ 1.66 \%  (3.87) & {\bf 91.55 } $\pm$ 0.00 \% &   98.94 \%\\
Wine &  76.14 $\pm$ 4.29 \%  &79.47 $\pm$ 4.25 \% (3.97)  &79.36  $\pm$ 5.07 \% (4.23) &  76.14   $\pm$ 0.00 \% & 100.00 \%\\
Image Segmentation  &  79.61 $\pm$ 4.43 \%  &75.60 $\pm$ 5.13 \% (4.57) &75.31 $\pm$ 4.97 \% (4.13) & {\bf 79.62 } $\pm$ 0.00 \%  &  98.48 \%\\
Letters Recognition   &  88.41 $\pm$ 1.34 \%  &87.00 $\pm$ 1.68 \% (3.80) &89.29  $\pm$ 0.00 \% (9.00) & {\bf 89.52 } $\pm$ 0.00 \% & 96.70 \% \\
\hline
\hline
  \end{tabular}
\caption{The average recognition rates of the ensembles of Parzen Windows classifiers selected by a GA with different objective functions. The MVE and ME average ensemble sizes are shown in parentheses.}   \label{t32}
\end{center}
\end{scriptsize}
\end{table}

In order to compare the performance of the classifier-free approach with the traditional classifier-based approach, we also evaluated the single GA search with MVE and with ME as the objective functions. For these two schemes, classifiers were constructed using given feature subset pools, and the GA search evaluated the results directly from the classifier outputs, regardless of the clustering diversities of their feature subsets. For MVE, the ensembles were selected for the minimum ensemble errors; and for ME, the ensembles were chosen for the minimum average of the individual classifier error. All classifiers were combined using MAJ as the fusion function.

For the single GA search, we set $32$ individuals in the population with $500$ generations. The mutation rate was set to $\frac {1} {L}$, where L is the length of the mutated binary string \cite{ei-pa-1998}, and the crossover probability was set to $50 \%$. These parameters were defined empirically. A threshold of $3$ classifiers was applied as the minimum number of classifiers for the EoC during the whole search. The experiments were repeated $30$ times for statistical evaluation.

We note that, in general, the MVE, and even the ME, have much better performances than all the other clustering diversity indices (Table \ref{t30} $\sim$ \ref{t32}). This is not surprising, since the clustering diversity indices do not take into account the classifier outputs. In our experiments, the ME does not converge to the minimum ensemble size, but we found that several ensembles can achieve the same ME, which explains why ME could have ensemble sizes that are larger than the minimum. This is reasonable, since the pool consists of only $10$ classifiers. Moreover, given that all GA searches with the clustering diversity indices converge to the minimum number of classifiers (fixed to 3 in our experiments), it is understandable that the single GA search with the clustering diversity indices performs less well.

Given that we are not only looking for the optimum performances from these clustering diversity indices, but also a pre-selection for the more refined ensemble selection methods, this premature convergence of the single GA is not desirable. In order to resolve the problem of convergence into the minimum ensemble size, we carried out a MOGA search in our next experiment. 

\subsubsection{Search with MOGA for the UCI Machine Learning Problems}

%\subsection{Search with the Multi-Objective Genetic Algorithm}

As we can observe from the single GA search, there is a technical problem with the use of pairwise diversity as an objective function: the search algorithm will converge to the minimum number of feature subsets (and hence the minimum size of the ensemble) with the maximum clustering diversity, which means that the search algorithm systematically prefers the smaller ensembles to bigger ones \cite{ko-co-2006}. It turns out that we will, in effect, encounter two problems if we use pairwise diversities. So, aside from optimizing the diversity, we should at the same time avoid minimizing the number of feature subsets.

Given the challenges posed by ensemble selection, the prospect of satisfying multi-objective problems makes the MOGA a desirable alternative. We thus define two objectives for the search: the optimization of diversity (and hence the minimization of two Wallace indices, the Fowlkes-Mallows index, the Rand index, the Jacard index and the maximization of Mirkin's Metric) and the maximization of the number of feature subsets. Although we only care about diversity, maximizing the number of feature subsets can prevent the search from converging to the minimal number of feature subsets (and hence the minimum size of the ensemble).

\begin{table}
\begin{scriptsize}
\begin{center}
  \begin{tabular}{|c|c|c|c|c|c|}
   \hline
   \hline
  &  Mirkin's & Wallace Index-1     & Wallace  Index-2     &  Fowlkes-Mallows  &    Rand  \\
       \hline
         \hline
  Pima-Diabetes  &   80.10  $\pm$ 2.03 \% & 77.87  $\pm$ 1.18 \% & 79.07  $\pm$ 2.56 \% & 79.96  $\pm$ 1.77 \% & 79.13  $\pm$ 1.90 \%    \\
Liver-Disorders &  72.78  $\pm$ 2.97 \% & 74.08  $\pm$ 2.83 \% & 74.26  $\pm$ 2.53 \% & 71.93  $\pm$ 3.54 \% & 72.94  $\pm$ 3.10 \%  \\
Wisconsin Breast-Cancer  &    92.28  $\pm$ 1.82 \% &{\bf  92.78 } $\pm$ 1.96 \% & 92.18  $\pm$ 1.26 \% & 92.30  $\pm$ 2.05 \% & 91.99  $\pm$ 2.01 \%  \\
Wine  &  74.47  $\pm$ 2.40 \% & 74.94  $\pm$ 2.30 \% & 74.33  $\pm$ 1.67 \% & 75.58  $\pm$ 3.51 \% & 75.44  $\pm$ 3.63 \%  \\
Image Segmentation  &   74.80  $\pm$ 5.08 \% & 75.47  $\pm$ 4.66 \% & 75.04  $\pm$ 3.60 \% & 75.72  $\pm$ 3.03 \% & 74.89  $\pm$ 3.68 \%  \\
Letters Recognition  & 79.13  $\pm$ 2.92 \% & 80.10  $\pm$ 2.74 \% & 80.45  $\pm$ 1.29 \% & 80.89  $\pm$ 1.48 \% & 78.98  $\pm$ 3.50 \% \\
 \hline
 \hline
 &  Jacard & M.V.E & M.E. & ALL  & Oracle \\
    \hline
         \hline
Pima-Diabetes & 79.91  $\pm$ 1.87 \% & 79.33  $\pm$ 2.12 \% & 79.48 $\pm$ 2.06 \% &{\bf  82.55 } $\pm$ 0.00 \% & 98.18 \%\\
Liver-Disorders  & 74.01  $\pm$ 2.47 \% & 74.07  $\pm$ 3.56 \% & 73.79 $\pm$ 2.92 \% & {\bf 76.39  } $\pm$ 0.00 \% & 100.00 \% \\
Wisconsin Breast-Cancer  & 88.87  $\pm$ 1.79 \% & 92.48  $\pm$ 0.95 \% & 92.46 $\pm$ 1.28 \% & 92.61  $\pm$ 0.00 \% & 99.65 \% \\
Wine & {\bf 76.29 } $\pm$ 3.04 \% & 75.51  $\pm$ 2.84 \% & 74.27 $\pm$ 2.74 \% & 76.14    $\pm$ 0.00 \% & 97.73 \%\\
Image Segmentation  & 75.55  $\pm$ 4.94 \% & 74.16  $\pm$ 3.67 \% & 74.11 $\pm$ 4.00 \% & {\bf 78.19}   $\pm$ 0.00 \% & 97.29 \%\\
Letters Recognition   & 80.10  $\pm$ 2.14 \% & 80.30  $\pm$ 2.29 \% & 77.59 $\pm$ 3.82 \% & {\bf 83.08}   $\pm$ 0.00 \%  &94.78 \% \\
\hline
\hline
  \end{tabular}
\caption{The average recognition rates of the ensembles of KNN classifiers selected by the MOGA with different objective functions on problems extracted from the UCI Machine Learning Repository.}   \label{t33}
\end{center}
\end{scriptsize}
\end{table}

\begin{table}
\begin{scriptsize}
\begin{center}
  \begin{tabular}{|c|c|c|c|c|c|}
   \hline
   \hline
  &  Mirkin's & Wallace  Index-1     & Wallace  Index-2     &  Fowlkes-Mallows   &    Rand   \\
       \hline
         \hline
Pima-Diabetes  &   4.33 & 4.27 & 4.33 & 5.00 & 4.02    \\
Liver-Disorders &   3.69 & 4.29 & 4.16 & 4.06 & 4.27  \\
Wisconsin Breast-Cancer  & 3.92 & 4.12 & 3.70 & 4.19 & 4.24    \\
Wine  &  4.47 & 4.28 & 3.66 & 4.47 & 3.93  \\
Image Segmentation  &  3.67 & 4.31 & 4.50 & 4.47 & 4.33 \\
Letters Recognition  &   4.00 & 4.00 & 4.31 & 4.47 & 3.67\\
      \hline
         \hline
&  Jacard & M.V.E & M.E. & ALL   & \\
    \hline
         \hline
Pima-Diabetes & 4.43 & 4.16 & 4.29  & 10.00 & \\
Liver-Disorders & 3.99 & 4.02 & 3.95 & 10.00  & \\
Wisconsin Breast-Cancer & 4.23 & 4.26 & 3.87 & 10.00 & \\
Wine  & 4.83 & 4.24 & 3.60  & 10.00 & \\
Image Segmentation  & 4.83 & 4.24 & 3.60  & 10.00 & \\
Letters Recognition  & 4.39 & 4.21 & 3.38 & 10.00 & \\
\hline
  \end{tabular}
\caption{The average ensemble sizes of KNN classifiers selected by the MOGA with different objective functions on problems extracted from the UCI Machine Learning Repository.}   \label{t34}
\end{center}
\end{scriptsize}
\end{table}

\begin{table}
\begin{scriptsize}
\begin{center}
  \begin{tabular}{|c|c|c|c|c|c|}
   \hline
   \hline
  &  Mirkin's & Wallace Index-1     & Wallace  Index-2     &  Fowlkes-Mallows  &    Rand  \\
       \hline
         \hline
  Pima-Diabetes  &   75.89 $\pm$ 2.62 \%  &75.08 $\pm$ 3.48 \%  & 76.03 $\pm$ 2.20 \%  & 74.97 $\pm$ 2.65 \%  & 74.69  $\pm$ 2.68 \%    \\
Liver-Disorders &  56.88 $\pm$ 2.50 \%  & 57.41 $\pm$ 2.31 \%  & 56.93 $\pm$ 2.24 \%  & 57.17 $\pm$ 3.13 \%  & 57.56   $\pm$ 3.06 \%   \\
Wisconsin Breast-Cancer  &   93.62 $\pm$ 2.01 \%  & 93.93 $\pm$ 1.65 \%  & {\bf 94.36} $\pm$ 1.43 \%  & 93.60 $\pm$ 2.01 \%  & 93.48   $\pm$ 1.69 \%    \\
Wine  &  95.81 $\pm$ 2.59 \%  & 96.20 $\pm$ 0.97 \%  & 92.74 $\pm$ 1.63 \%  & 95.27 $\pm$ 2.44 \%  & 95.61   $\pm$ 1.93  \%     \\
Image Segmentation  &   50.67 $\pm$ 23.37 \%  & 57.84 $\pm$ 15.54 \%  & 63.78 $\pm$ 13.54 \%  & 61.60 $\pm$ 13.05  \%  & 64.78  $\pm$ 15.23 \%  \\
Letters Recognition  &  80.79 $\pm$ 2.41 \%  & 81.85 $\pm$ 2.10 \%  & 82.10 $\pm$ 1.78 \%  & 81.98 $\pm$ 1.19 \%  & 81.16  $\pm$ 1.60 \%   \\
 \hline
 \hline
 &  Jacard & M.V.E & M.E. & ALL  & Oracle \\
    \hline
         \hline
Pima-Diabetes & 75.68 $\pm$ 2.07 \%  & 75.62 $\pm$ 2.68 \%  & 74.58 $\pm$ 2.56 \%  & {\bf 77.86}  $\pm$  \% 0.00 & 93.23 \%\\
Liver-Disorders  &  56.77 $\pm$ 2.38 \%  & 56.53 $\pm$ 2.32 \%  & 57.46 $\pm$ 2.33 \%  &  {\bf 57.64}  $\pm$  \% 0.00 &88.19 \% \\
Wisconsin Breast-Cancer  & 91.46 $\pm$ 1.41 \%  & 94.02 $\pm$ 1.70 \%  & 93.67$\pm$ 1.81 \%  & 93.66  $\pm$ 0.00  \%  & 99.65 \%\\
Wine & 95.48 $\pm$ 1.11 \%  &  95.14 $\pm$ 2.86 \%  & 95.11$\pm$ 2.10 \%  & {\bf 96.59} $\pm$  \% 0.00 &100.00 \% \\
Image Segmentation  &     52.20 $\pm$ 18.43 \%  & 59.11 $\pm$ 12.58 \%  & 57.20 $\pm$ 11.25 \%  & {\bf 78.24 } $\pm$  \% 0.00 & 95.29 \%\\
Letters Recognition   &    81.76 $\pm$ 2.06 \%  & 81.50 $\pm$ 1.67 \%  & 81.27$\pm$ 1.80 \%  & {\bf 84.36 } $\pm$  \% 0.00 & 93.40 \%\\
\hline
\hline
  \end{tabular}
\caption{The average recognition rates of the ensembles of QDC classifiers selected by the MOGA with different objective functions on problems extracted from the UCI Machine Learning Repository.}  \label{t35}
\end{center}
\end{scriptsize}
\end{table}

\begin{table}
\begin{scriptsize}
\begin{center}
  \begin{tabular}{|c|c|c|c|c|c|}
   \hline
   \hline
  &  Mirkin's & Wallace  Index-1     & Wallace  Index-2     &  Fowlkes-Mallows   &    Rand   \\
       \hline
         \hline
Pima-Diabetes  &   4.31 &   4.12  &  4.49  &  4.30  &  3.94 \\
Liver-Disorders &   3.86  &  4.13 &   4.02  &  4.62  &  3.90      \\
Wisconsin Breast-Cancer  &   3.92 &  4.15 &   3.57   & 3.94 &   4.10     \\
Wine  &  4.35  &  4.22  &  3.85  &  4.29  &  3.78      \\
Image Segmentation  & 3.16  &  4.41  &  4.50  &  4.25  &  4.55  \\
Letters Recognition  &   3.79 &   4.08   & 4.61   & 4.62  &  3.84    \\
      \hline
         \hline
&  Jacard & M.V.E & M.E. & ALL   & \\
    \hline
         \hline
Pima-Diabetes &     4.42 &  4.16  &  4.56  &   10.00  & \\
Liver-Disorders & 4.19  &  4.38 &   3.93  & 10.00  & \\
Wisconsin Breast-Cancer &  4.20 &   3.81  &  4.11 & 10.00  & \\
Wine  & 4.53 &   4.35  &  3.95  & 10.00   & \\
Image Segmentation  &    3.48  &  3.81  &  3.72 &  10.00  & \\
Letters Recognition  & 4.43  &  4.14  &  3.81 &  10.00  & \\
\hline
  \end{tabular}
\caption{The average ensemble sizes of QDC classifiers selected by the MOGA with different objective functions on problems extracted from the UCI Machine Learning Repository.}   \label{t36}
\end{center}
\end{scriptsize}
\end{table}

\begin{table}
\begin{scriptsize}
\begin{center}
  \begin{tabular}{|c|c|c|c|c|c|}
   \hline
   \hline
  &  Mirkin's & Wallace Index-1     & Wallace  Index-2     &  Fowlkes-Mallows  &    Rand  \\
       \hline
         \hline
  Pima-Diabetes  & {\bf  78.49 }$\pm$ 1.56 \%  &75.00 $\pm$ 1.14 \%  &77.12 $\pm$ 2.58 \%  &78.18 $\pm$ 1.13 \%  &77.73   $\pm$ 2.02 \%    \\
Liver-Disorders &  68.66 $\pm$ 3.15 \%  &68.18 $\pm$ 3.52 \%  & 68.29 $\pm$ 4.39 \%  & 67.77 $\pm$ 3.90 \%  &67.55 $\pm$ 4.23 \%  \\
Wisconsin Breast-Cancer  &  90.83 $\pm$ 1.22 \%  &90.98 $\pm$ 1.08 \%  &90.86 $\pm$ 1.03 \%  &91.16 $\pm$ 1.22 \%  &90.25 $\pm$ 1.48 \%  \\
Wine  & 76.52 $\pm$ 1.61 \%  &79.06 $\pm$ 4.43 \%  &79.96 $\pm$ 1.35 \%  &78.60 $\pm$ 4.51 \%  &79.62   $\pm$ 5.08 \%    \\
Image Segmentation  & 75.53 $\pm$ 5.62 \%  &75.74 $\pm$ 5.42 \%  &76.33 $\pm$ 5.24 \%  &76.61 $\pm$3.28  \%  &75.79  $\pm$ 5.10 \%  \\
Letters Recognition  &  86.88 $\pm$ 2.13 \%  & 87.39 $\pm$ 1.96  \%  &87.70 $\pm$ 1.03 \%  &87.74 $\pm$ 1.14 \%  &86.83  $\pm$  2.06\%  \\
 \hline
 \hline
 &  Jacard & M.V.E & M.E. & ALL  &Oracle  \\
    \hline
         \hline
Pima-Diabetes &  77.57 $\pm$ 2.33 \%  &76.45 $\pm$ 2.78 \%  &77.62 $\pm$ 1.92 \%  &78.12   $\pm$ 0.00 \%  & 92.19 \%\\
Liver-Disorders  &  68.11 $\pm$ 3.55 \%  &68.23 $\pm$ 2.96 \%  &68.39$\pm$ 3.50 \%  &{\bf 70.83}  $\pm$ 0.00 \%  & 89.58 \% \\
Wisconsin Breast-Cancer  &   88.23 $\pm$ 1.47 \%  &91.27 $\pm$ 1.30 \%  &90.89    $\pm$ 1.34 \%  &{\bf 91.55 } $\pm$ 0.00 \%   & 98.94 \%\\
Wine &  78.66 $\pm$ 4.32 \%  &78.45 $\pm$ 4.10 \%  &{\bf 80.02}$\pm$ 4.29 \%  &76.14   $\pm$ 0.00 \%  & 100.00 \%\\
Image Segmentation  &  77.63 $\pm$ 5.86 \%  &75.94 $\pm$ 4.13 \%  &76.83$\pm$ 4.71 \%  &{\bf 79.62 } $\pm$ 0.00 \%  & 98.48 \%\\
Letters Recognition   &   87.46 $\pm$  1.49\%  &87.26 $\pm$ 1.61 \%  &87.45   $\pm$ 1.01 \%  &{\bf 89.52 } $\pm$ 0.00 \%  &96.70 \% \\
\hline
\hline
  \end{tabular}
\caption{The average recognition rates of the ensembles of PARZEN WINDOWS classifiers selected by the MOGA with different objective functions on problems extracted from the UCI Machine Learning Repository.}   \label{t37}
\end{center}
\end{scriptsize}
\end{table}

\begin{table}
\begin{scriptsize}
\begin{center}
  \begin{tabular}{|c|c|c|c|c|c|}
   \hline
   \hline
  &  Mirkin's & Wallace  Index-1     & Wallace  Index-2     &  Fowlkes-Mallows   &    Rand   \\
       \hline
         \hline
Pima-Diabetes  &  4.48  &  3.75 &   4.42  &  4.89 &   4.09     \\
Liver-Disorders &   3.98 &   4.30  &  4.11 &   4.45 &   3.84     \\
Wisconsin Breast-Cancer  &  4.06 &   4.17  &  3.65  &  4.19 &   4.10   \\
Wine  & 4.58  &  4.17  &  3.80 &   4.21   & 3.86     \\
Image Segmentation  & 3.41 &   4.32   & 4.44  &  4.46   & 4.70  \\
Letters Recognition  &  4.11  &  3.95  &  4.28 &   4.11 &   3.93     \\
      \hline
         \hline
&  Jacard & M.V.E & M.E. & ALL   & \\
    \hline
         \hline
Pima-Diabetes &    4.18   & 4.05 &   4.13 &  10.00 & \\
Liver-Disorders &   4.10  &  5.02  &  4.06  &   10.00 & \\
Wisconsin Breast-Cancer &   4.34  &  3.97 &  4.03  &  10.00 & \\
Wine  &  4.71 &   3.78   & 4.02  &  10.00 & \\
Image Segmentation  &    4.23 &   3.93  &  4.48  &  10.00 & \\
Letters Recognition  & 4.23  &  4.31  &  4.19 &  10.00 & \\
\hline
  \end{tabular}
\caption{The average ensemble sizes of Parzen Windows classifiers selected by the MOGA with different objective functions on problems extracted from the UCI Machine Learning Repository.}   \label{t38}
\end{center}
\end{scriptsize}
\end{table}

Like the experiments with the GA, we also used $32$ individuals in the population and $500$ generations here. The mutation rate was set to $\frac {1} {L}$, where L is the length of the mutated binary string \cite{ei-pa-1998}, and the crossover probability was set to $50 \%$. For both classifier-free ensemble selection (or feature subset selection) and classifier-based ensemble selection, a threshold of $3$ feature subsets or classifiers was applied as the minimum number of feature subsets or classifiers, and the experiments were repeated $30$ times.

Note that the MOGA solutions are non dominated (known as Pareto-optimal) solutions. In order to approach these solutions, we applied a non dominated sorting GA (NSGA2), developed by Deb \cite{de-mu-2002}. NSGA2 maintains the dual objective of the MOGA by using a fitness assignment scheme, which prefers non dominated solutions, and a crowded distance strategy, which preserves diversity among the solutions of each non dominated front.

First, we note that the MOGA search based on clustering diversity indices gives a larger population than the single GA does for classifier-free ensemble selection (Tables \ref{t34}, \ref{t36}, \ref{t38}). Although their population sizes are larger, the feature subsets selected with the MOGA generally, but not always, perform better than those selected with the single GA (Table \ref{t39}).

\begin{table}
\begin{scriptsize}
\begin{center}
  \begin{tabular}{|c|c|c|c|c|c|c|}
   \hline
     &  Pima & Liver & Wisconson &  Wine     & Image     &  Letter   \\
     &  -Diabete & -Disorder & Breast Cancer &       & Segmentation     &  Recognition   \\
       \hline
         \hline
KNN  &  1e-06   & 1e-07    & 2e-09    &  8e-04  & 2e-09   &  6e-04  \\
QDC  &  2e-09   & 0.0829  &  0.2513  &  2e-09 & 1e-09   &  2e-09   \\
PWC  &  2e-09   & 0.3482  &  0.1891   &  8e-04 &  2e-09  &   2e-09  \\
\hline
  \end{tabular}
\caption{The $p$ value of the recognition rates between the classifier-free MOGA search and the classifier GA search.}   \label{t39}
\end{center}
\end{scriptsize}
\end{table}

By contrast, the MOGA search based on ME or MVE does not perform better than the single GA search for classifier-based ensemble selection. This is understandable, since both ME and MVE benefit directly from the classifier outputs, with the result that the maximum ensemble size does not help much in improving the results.

Interestingly, we observe that, with the MOGA search, most objective functions, including clustering diversities for classifier-free ensemble selection and ME and MVE for classifier-based ensemble selection, gave similar performances  (Table \ref{t33}, \ref{t35}, \ref{t37}). The reasonably small standard deviations indicate that their performances are quite stable in different replications. There seems to be no index that is apparently best for both classifier-free ensemble selection and classifier-based ensemble selection. The best solutions seem to be problem-dependent. According to the 'no free lunch' theorem \cite{wh-re-2000, wo-no-1997}, there is no single search algorithm that will always be the best for all problems. This phenomenon can be observed in our experiments.

Although the experiments on UCI machine learning problems suggest that a classifier-free ensemble selection scheme might be applicable, these experiments were carried out on small databases (apart from the letter recognition problem, where the number of samples $\le$ 3000) with a small number of features (apart from the breast cancer problem, where the number of features was $\le$ 20) and relatively small pools (total classifiers $= 10$). In other words, we knew that clustering diversity might work in classifier-free ensemble selection, but only for small-scale problems.

\subsection{Evaluation on the $NIST SD19$ Handwritten Numeral Database}

Although the experiments on the UCI machine learning problems suggest that a classifier-free ensemble selection scheme might be applicable, these experiments were carried out on small databases (apart from the letter recognition problem, where the number of samples $\le$ 3000) with a small number of features (apart from the breast cancer problem, where the number of features $\le$ 20) and relatively small pools (total classifiers $= 10$). In other words, we knew that clustering diversity might work in classifier-free ensemble selection, but only for small-scale problems.

We wanted to know whether or not classifier-free ensemble selection would be applicable in a large-scale problem. Similar to the experiments on problems extracted from the UCI data repository, these experiments were executed with both the single GA search and the MOGA search.

The experiments were performed on a $10$-class handwritten-numeral problem. The data were extracted from $NIST SD19$, essentially as in \cite{tr-op-2004}. We first defined $100$ feature subspaces for classifier-free ensemble selection (or feature subset selection), each feature subspace containing $32$ features extracted from the total of $132$ features. For classifier-based ensemble selection, these $100$ feature subspaces were used to train $100$ corresponding KNN classifiers. We used nearest neighbor classifiers ($K=1$) for the KNN classifiers.

\begin{figure}[!htb]
      \begin{center}
      \includegraphics[width=.75\textwidth]{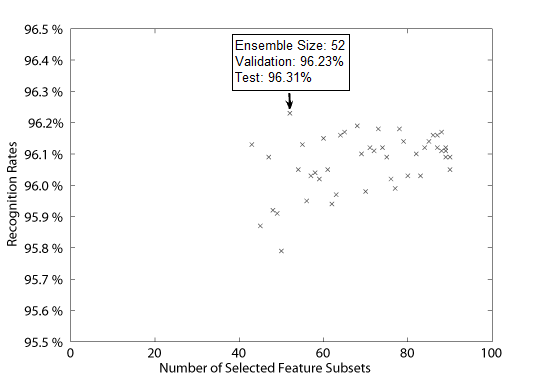}
      \caption{The validated recognition rates of individuals on the Pareto front.  E.S. = Ensemble Size; V.R.R. = Validation Recognition Rate in percent.}
\label{rec}
      \end{center}
    \end{figure}

Several databases were used:

\begin{itemize}

\item Training set: \\
This set, containing $5000$ data points ($NIST SD19$ $hsf\_\{0-3\}$), was used to create $100$ KNN in Random Subspaces for classifier-based ensemble selection. Note that, since classifier-free ensemble selection does not require classifiers, this set was not used for classifier-free ensemble selection until the final evaluation stage. Also note that this set is used only for the KNN classifiers and not for search purposes.

\item Optimization set: \\
This set, containing $10000$ data points ($NIST SD19$ $hsf\_\{0-3\}$), was used for the GA and the MOGA search for both classifier-free ensemble selection and classifier-based ensemble selection. In the case of classifier-free ensemble selection, we measured the clustering diversities of various combinations of feature subsets, and, in the case of classifier-based ensemble selection, we measured the ME and MVE of various ensembles of classifiers.

For both the GA and MOGA search algorithms, we set at $128$ the number of individuals in the population and $500$ generations, which means that $64,000$ ensembles were evaluated in each experiment. The mutation rate was set to $\frac {1} {L}$, where $L$ is the length of the mutated binary string  \cite{ei-pa-1998}, and the crossover probability was set to $50 \%$. During the whole search, a threshold of $3$ feature subsets or classifiers was applied as the minimum number of feature subsets or classifiers for both classifier-free ensemble selection and classifier-based ensemble selection. All the experiments were carried out with $8$ different objective functions ($6$ clustering diversity measures for classifier-free ensemble selection, ME and MVE for classifier-based ensemble selection) and $30$ replications.

\item Validation set: \\
This set, containing $10000$ data points ($NIST SD19$ $hsf\_\{0-3\}$), was used to evaluate all the individuals according to the defined objective function, and then to store those individuals in a separate archive after each generation \cite{ra-an-2006} (see Fig. \ref{arch}) for both classifier-free ensemble selection and classifier-based ensemble selection. Note that the archive mechanism is designed to avoid overfitting of the defined objective functions, and has been shown to be capable of doing so \cite {ra-an-2006}, and that these objective functions may or may not represent classification accuracy. Moreover, at this stage, there are no classifiers for classifier-free ensemble selection.

For classifier-free ensemble selection, the objective functions are clustering diversities, and so we evaluated them on the validation set and stored the individuals of its Pareto front in a separate archive. For classifier-based ensemble selection, the objective functions are ME and MVE, and thus we evaluated ensemble performances using ME or MVE as fusion functions on the validation set and stored their Pareto front in an archive.

The validation set was also used for the final evaluation of the classifier-free MOGA search. Since this search gives a group of solutions, and because each solution is an ensemble of feature subsets, it is difficult to say which solution will be the best in terms of recognition rate. As a result, these solutions of combinations of feature subsets need to be further evaluated. To do so, we would need to construct EoCs based on the groups of feature subspaces found, and then evaluate the performances of these ensembles (Fig. \ref{pareto} \& Fig. \ref{rec}). The solutions stored in the archive were used to construct ensembles using the training set, and their performances evaluated on the validation set. The best solution found on the validation set was then evaluated on the test set.

\item Test set: \\
This set, containing $60089$ data points ($NIST SD19$ $hsf\_\{7\}$), was used to evaluate the ensembles selected by classifier-free ensemble selection and by classifier-based ensemble selection. A MAJ is used as the fusion function for classifier combination, because of its stable performance as reported in the literature \cite{ru-cl-2005}.
\end{itemize}

Note that, in accordance with the definition of the validation set, we used the global validation of all solutions for each generation, and maintained the best solutions in an external archive. The best solution defined in terms of ME in the Pareto front was selected, and its performance evaluated on the test set.

\begin{table}
\begin{scriptsize}
\begin{center}
\begin{tabular}{|c|}
\hline
\hline
 ALL\\
\hline
96.28 \% (100.00) \\
\hline
\hline
\end{tabular}  \\
\bigskip
Classifier-Based Ensemble Selection \\
\bigskip
\begin{tabular}{|c|c|}
\hline
\hline
ME & MVE  \\
\hline
94.18 $\pm$ 0.00\% (3.00 $\pm$ 0.00) & {\bf 96.45} $\pm$ 0.05\% (24.53 $\pm$ 3.58) \\
\hline
\hline
\end{tabular}  \\
\bigskip
Classifier-Free Ensemble Selection \\
\bigskip
\begin{tabular}{|c|c|c|}
\hline
\hline
 Wallace Index-1 & Wallace Index-2 & Fowlkes-Mallows    \\
\hline
92.55 $\pm$ 0.55\%  ((3.00 $\pm$ 0.00) & 92.61 $\pm$ 0.43 \%  (3.00 $\pm$ 0.00) & 93.06 $\pm$ 0.14\%  (3.00 $\pm$ 0.00)\\
\hline
\hline
Rand & Jacard &  Mirkin's \\
\hline
92.25 $\pm$ 0.56\%  (3.00 $\pm$ 0.00)  &  92.22 $\pm$ 0.10\%  (3.00 $\pm$ 0.00) &   93.03 $\pm$ 0.50\%  (3.00 $\pm$ 0.00) \\
\hline
\hline
\end{tabular}
\caption{The average recognition rates on the test data of ensembles searched by a GA with different objective functions, including: original clustering diversity measures, compared with MEs and MVEs. Simple majority voting was used as the fusion functions, and the ensemble sizes are indicated in parentheses.}   \label{t40}
\end{center}
\end{scriptsize}
\end{table}

Note that, in accordance with the definition of the validation set, we used the global validation of all solutions for each generation, and maintained the best solutions in an external archive. The best solution defined in terms of ME in the Pareto front was selected, and its performance evaluated on the test set. 

\subsubsection{Search with Single GA for the Handwritten Numeral Recognition}

%\subsection{Search with Single GA for Handwritten Numeral Recognition}

    \begin{figure}[!htb]
      \begin{center}
      \includegraphics[width=.75\textwidth]{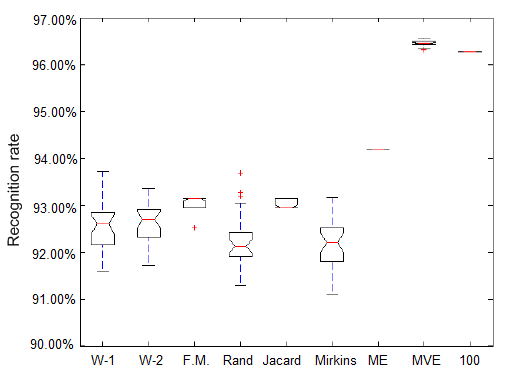}
      \caption{The average recognition rates achieved by EoCs selected by modified clustering diversities with the single GA, compared with Mean Classifier Error (ME), Majority Voting Error (MVE), and the ensemble of all (100) KNN classifiers.}
\label{box_GA}
      \end{center}
    \end{figure}

    \begin{figure}[!htb]
      \begin{center}
      \includegraphics[width=.75\textwidth]{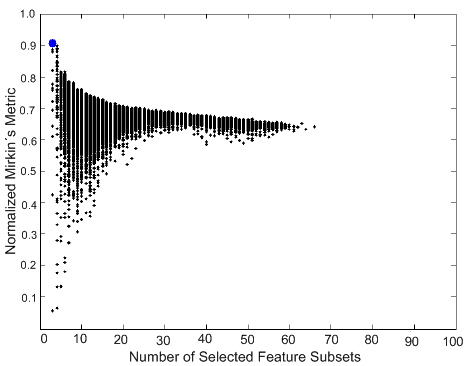}
      \caption{The evaluated population (diamonds) and selected solution (the circle) based on the single GA search with Mirkin's Metric as the objective function. The number of selected feature subsets is shown to illustrate the process of convergence into the minimum feature subset size.}
\label{size_GA}
      \end{center}
    \end{figure}

We performed a number of experiments directly, using the various objective functions for ensemble selection that had been evaluated by the GA search. We tested $6$ clustering diversity measures for classifier-free ensemble selection (or feature subset selection), and ME and MVE for classifier-based ensemble selection. We then compared the performances of the EoCs selected by the two selection methods.

For classifier-based ensemble selection, the EoCs selected by MVE achieved an average $96.45 \%$ classification accuracy, while those selected by ME had only a $94.18 \%$ recognition rate (Table \ref{t40}; Fig. \ref{box_GA}). Note that the EoCs found by MVE have, in general,$19 \sim 35$ classifiers. However, for classifier-free ensemble selection, the GA search led to the minimum number of feature subsets (Fig. \ref{size_GA}). Nevertheless, there is a huge gap between the performances of classifier-free ensemble selection using clustering diversity indices and those of classifier-based ensemble selection using MVE. We note that even classifier-based ensemble selection using simple ME can perform better than classifier-free ensemble selection using clustering diversity measures as objective functions.

However, this does not mean that the idea of classifier-free ensemble selection is not a valid one. As we have already stated, the major problem with the GA search is its convergence to the minimum feature subset size ($3$ feature subsets), and thus the problem resides more in the search algorithm than in the choice of objective functions. That is why we applied the MOGA for classifier-free ensemble selection.

\subsubsection{Search with MOGA for the Handwritten Numeral Recognition}
%\subsubsection {Search with MOGA for the Handwritten Numeral Recognition}

%\subsection{Multi-Objective Genetic Algorithm for Ensemble Selection for Handwritten Numeral Recognition}

\begin{table}
\begin{scriptsize}
\begin{center}
\begin{tabular}{|c|}
\hline
\hline
 ALL\\
\hline
96.28 \% (100.00) \\
\hline
\hline
\end{tabular} \\
\bigskip
Classifier-Based Ensemble Selection \\
\bigskip
\begin{tabular}{|c|c|}
\hline
\hline
ME & MVE  \\
\hline
96.26 $\pm$ 0.08\% (48.83 $\pm$ 5.75) & 96.25 $\pm$ 0.04\% (49.25 $\pm$ 5.59) \\
\hline
\hline
\end{tabular} \\
\bigskip
Classifier-Free Ensemble Selection \\
\bigskip
\begin{tabular}{|c|c|c|}
\hline
\hline
 Wallace Index-1 & Wallace Index-2 & Fowlkes-Mallows    \\
\hline
96.24 $\pm$ 0.08\% (50.88 $\pm$ 5.34) & 96.25 $\pm$ 0.06 \% (51.08 $\pm$ 4.46) & 96.25 $\pm$ 0.08\% (50.42 $\pm$ 4.93)   \\
\hline
\hline
  Rand & Jacard &  Mirkin's \\
\hline
96.23 $\pm$ 0.08\% (51.95 $\pm$ 4.09) & 96.26 $\pm$ 0.06\% (52.91 $\pm$ 4.63) & 96.19 $\pm$ 0.08\% (50.75 $\pm$ 4.61) \\
\hline
\hline
\end{tabular}
\caption{The average recognition rates on the test data of ensembles searched by a MOGA with different objective functions, including: original clustering diversity measures, three approximations of classifier diversity measures, compared with MEs and MVEs. Simple majority voting was used as the fusion functions, and the ensemble sizes are indicated in parentheses.}   \label{t41}
\end{center}
\end{scriptsize}
\end{table}

For classifier-free ensemble selection, the use of the MOGA search emphasizes the optimization of the clustering indices, as well as the maximization of the number of feature subsets. While the latter is no less relevant to better ensemble performance, it does avoid the problem of minimum ensemble size convergence that occurred in the GA search. While a MOGA search might not be necessary for classifier-based ensemble selection, we performed one nonetheless, so that we could compare the results of classifier-based ensemble selection with those of classifier-free ensemble selection.

{figure}[!htb]
   \begin{figure}[!htb]
      \begin{center}
      \includegraphics[width=.75\textwidth]{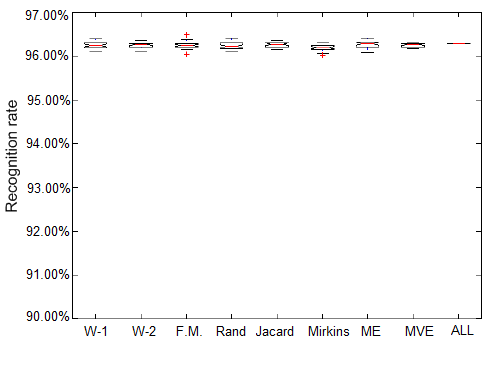}
      \caption{Box plot of the classifier-free ensemble selection schemes using a MOGA compared with the classifier-based ensemble selection using MEs and MVEs as objective functions.}
\label{box_MOGA}
      \end{center}
    \end{figure}

 \begin{figure}[!htb]
      \begin{center}
      \includegraphics[width=.85\textwidth]{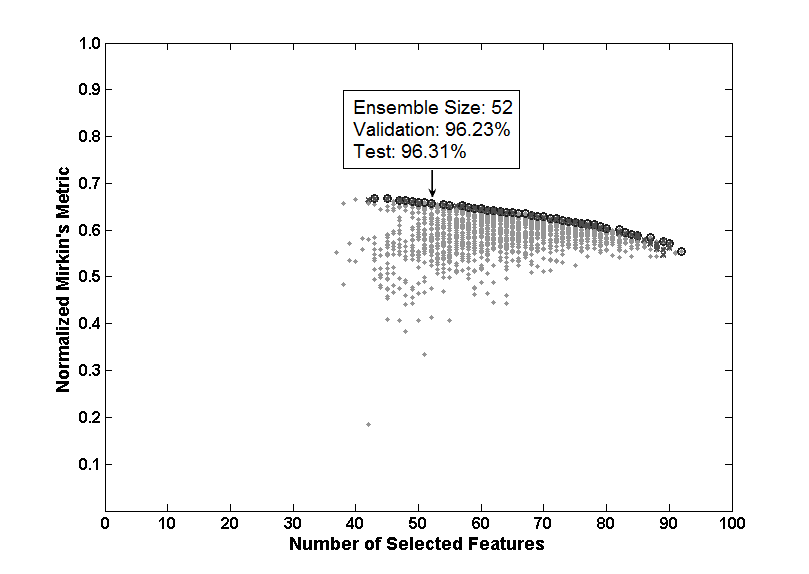}
      \caption{The Pareto front of the MOGA search  for the classifier-free ensemble  selection scheme. The evaluated population (diamonds), the population in the Pareto front (circles) and the validated solution (crosses) based on the MOGA search with Mirkin's Metric and the number of selected feature subsets the objective functions.  The best performance evaluated on the validation set is shown in the text boxes.}
\label{pareto}
      \end{center}
    \end{figure}{figure}[!htb]

\begin{table}
\begin{scriptsize}
\begin{center}
  \begin{tabular}{|c|c|c|c|c|c|c|c|}
   \hline
   \hline
  Mirkin's & Wallace  Index-1     & Wallace  Index-2     &  Fowlkes-Mallows   &    Rand  &  Jacard & M.V.E & M.E.  \\
 \hline
0.0001  &  0.2005  &  0.2005  &  0.0428  &  0.2005  &  0.5847 &   0.8555  &  0.0161 \\
\hline
  \end{tabular}
\caption{The p-value of the hypothesis test on the recognition rates of ensembles selected by various objective functions compared with that of the ensemble of all classifiers.}  \label{t42}
\end{center}
\end{scriptsize}
\end{table}

First, we note that, because we used a MOGA, classifier-free ensemble selection with clustering diversity indices no longer converged to $3$ feature subsets  (Fig. \ref{pareto}). In general, the population selected from the Pareto front has about half the feature subsets of the total pool (see Table \ref{t41}). This could allow further, more refined ensemble selection.

Moreover, we note that, in general, the feature subsets selected by classifier-free ensemble selection with clustering diversity indices construct adequate ensembles. The recognition rates achieved by these ensembles are very close to those achieved when all the classifiers are used (Fig. \ref{box_MOGA}).  In fact, the significances are usually $p \ge 0.01$ (Table \ref{t42}).

For classifier-based ensemble selection, ME also benefits from the MOGA scheme, and even slightly outperforms MVE as an objective function in a MOGA (See Table \ref{t41}). By contrast, MVE did not perform quite as well as in a single GA, but the difference is rather small ($0.20 \%$). With a MOGA, MVE selected $49.25$ classifiers on average, many more than it did with the simple GA.

The results of using the clustering diversities in classifier-free ensemble selection are encouraging, and all of them performed as well as the ensemble of all classifiers, but the ensemble sizes were cut in half. Furthermore, there is no clear difference among the various clustering diversity measures (Fig. \ref{box_MOGA}). This indicates that data diversity can be used to carry out ensemble selection in Random Subspaces, and that the proposed classifier-free ensemble selection scheme using clustering diversity measures as objective functions does work.

\subsection{Classifier-Free Ensemble Selection Combined with Pairwise Fusion Functions for Handwritten Numeral Recognition}
While MAJ is one of the fusion functions most often used for combining classifiers, it is not necessarily the optimum choice. In our experiment on handwritten numeral recognition, in which all the ensembles were combined with MAJ, classifier-based ensemble selection using MVE as the objective function, which uses MAJ to evaluate the ensembles, performed better than classifier-free ensemble selection using clustering diversity as the objective function.

However, if we apply other fusion functions - such as the pairwise fusion matrix with the majority voting rule (PFM-MAJ)  \cite{ko-ev-2006, ko-pa-2007} - classifier-based ensemble selection using MVE might not be the best scheme. It turns out that the performances of ensembles selected by classifier-free ensemble selection can be further improved by using better fusion functions. As we can see in Table \ref{t43}, the recognition rates of ensembles applying PFM-MAJ are apparently better than those applying the simple MAJ.

Moreover, for the MOGA search, when PFM-MAJ was used as the fusion function, classifier-free ensemble selection using the clustering diversity indices outperformed the classifier-based ensemble selection using MVE.

\begin{table}
\begin{scriptsize}
\begin{center}
\begin{tabular}{|c|}
\hline
\hline
 ALL\\
\hline
96.28 \% (100.00)\\
\hline
\hline
\end{tabular}  \\
\bigskip
Classifier-Based Ensemble Selection \\
\bigskip
\begin{tabular}{|c|c|}
\hline
\hline
ME & MVE  \\
\hline
96.89 $\pm$ 0.05\%  (48.83 $\pm$ 5.75) & 96.78 $\pm$ 0.09\%  (49.25 $\pm$ 5.59)  \\
\hline
\hline
\end{tabular}  \\
\bigskip
Classifier-Free Ensemble Selection \\
\bigskip
\begin{tabular}{|c|c|c|}
\hline
\hline
 Wallace Index-1 & Wallace Index-2 & Fowlkes-Mallows    \\
\hline
96.91 $\pm$ 0.05\% (50.88 $\pm$ 5.34) & 96.90 $\pm$ 0.04 \%  (51.08 $\pm$ 4.46) & 96.90 $\pm$ 0.04\%  (50.42 $\pm$ 4.93)  \\

\hline
\hline
 Rand &  Jacard & Mirkin's \\
\hline
  96.90 $\pm$ 0.04\%   (51.95 $\pm$ 4.09)  &  96.89 $\pm$ 0.03\% (52.91 $\pm$ 4.63) & 96.88 $\pm$ 0.08\% (50.75 $\pm$ 4.61)\\
\hline
\hline
\end{tabular}
\caption{The average recognition rates on the test data of ensembles searched by MOGA with different objective functions. The pairwise confusion matrix applying the pairwise-majority voting was used as the fusion function. The ensemble sizes are the same as those in Table. \ref{t41}.}   \label{t43}
\end{center}
\end{scriptsize}
\end{table}

\section{Discussion}
In this paper, we examined whether or not clustering diversity can represent the data diversity of different feature subsets in Random Subspaces, and whether or not the use of clustering diversity as the data diversity measure could allow us to apply a classifier-free ensemble selection scheme.

First, for classifier-free ensemble selection, we used the single GA as the search algorithm. We found that, with the clustering diversity indices as objective functions, it tends to converge to the minimum number of feature subsets, which makes a classifier-free ensemble selection scheme less useful.

Then, in order to compensate for the problem of minimum feature subset convergence of the clustering diversities, we used the MOGA as the search algorithm. The clustering diversity measures yielded encouraging performances as objective functions for the classifier-free ensemble selection scheme.

The only major cost is the evaluation of the solutions found on the Pareto front after the MOGA search. This requires the training of a classifier for each feature subset selected to evaluate the performances of ensembles, so that the best ensemble can be chosen. Compared with a traditional ensemble selection scheme, which requires the training of all classifiers and combinations of all the ensembles evaluated, the proposed scheme offers an interesting alternative. This approach will be especially attractive for tackling problems with a large classifier pool and time-consuming classifier training.

\section{Conclusion}
In this paper, we argue that clustering diversities actually represent the data diversities of the various feature subsets in the Random Subspaces ensemble creation method. These data diversities can be measured with the help of clustering diversities without any classifier training. As a result, the feature subsets can be selected by clustering diversities to construct the classifiers in Random Subspaces.

Applying the MOGA search, we show that the ensembles selected by the clustering diversities had performances comparable to those selected by MVE, which is regarded as one of the best objective functions for ensemble selection \cite {ru-cl-2005}. The results are encouraging. Based on our exploratory work, we have drawn up some implications for the classifier-free ensemble selection approach:
\begin{enumerate}
\item In Random Subspaces, with the MOGA search the clustering diversity measures are good objective functions for ensemble selection.
\item In Random Subspaces, the ensembles selected by the different clustering diversity measures have so far been found to have similar performances based on the MOGA search.
\item Using clustering partition diversity measures as objective functions for feature subset selection, the MOGA search may be more pertinent than the GA search.
\end{enumerate}

Even though the clustering partition diversities might only be able to represent data diversities in Random Subspaces, there is still no adequate measure for data diversities for Bagging and Boosting. It will be of great interest to figure out how to measure the data diversities for other ensemble generation methods.

\section*{Acknowledgment}
This work was supported in part by grant OGP0106456 to Robert Sabourin from the NSERC of Canada.

\end{document}